\documentclass[10pt,twocolumn,letterpaper]{article}

\usepackage{cvpr}              %

\usepackage[dvipsnames]{xcolor}

\usepackage{wrapfig,lipsum}
\usepackage[utf8]{inputenc} %
\usepackage[T1]{fontenc}    %
\usepackage{url}            %
\usepackage{booktabs}       %
\usepackage{amsfonts}       %
\usepackage{nicefrac}       %
\usepackage{microtype}      %
\usepackage{multirow}
\usepackage{graphicx}
\usepackage{subcaption}
\usepackage{times}
\usepackage{makecell}
\newcommand{\myparagraph}[1]{\noindent\textbf{#1}}
\newcommand{\suppl}{Supplement}
\usepackage{colortbl}
\definecolor{lightgray}{gray}{0.88}

\usepackage{amsmath,amsfonts,bm}

\def\eqref#1{equation~\ref{#1}}

\def\1{\bm{1}}

\def\mS{{\bm{S}}}

\DeclareMathAlphabet{\mathsfit}{\encodingdefault}{\sfdefault}{m}{sl}
\SetMathAlphabet{\mathsfit}{bold}{\encodingdefault}{\sfdefault}{bx}{n}

\DeclareMathOperator*{\argmin}{arg\,min}

\newcommand{\ours}{{SPU }}

\usepackage[accsupp]{axessibility}
\definecolor{cvprblue}{rgb}{0.21,0.49,0.74}
\usepackage[pagebackref,breaklinks,colorlinks,citecolor=cvprblue]{hyperref}

\def\paperID{16778} %
\def\confName{CVPR}
\def\confYear{2024}

\title{Overcoming Generic Knowledge Loss with Selective Parameter Update}

\author{Wenxuan Zhang\thanks{Email: \texttt{wenxuan.zhang@kaust.edu.sa}}
\,\,
Paul Janson$^{1,2}$\thanks{Work done during internship at KAUST}
\,\,
Rahaf Aljundi$^3$
\,\,
Mohamed Elhoseiny$^1$\\
$^1$KAUST \,
$^2$Concordia University \,
$^3$Toyota Motor Europe \\
}

\begin{document}
\maketitle
\begin{abstract}
Foundation models encompass an extensive knowledge base and offer remarkable transferability. However, this knowledge becomes outdated or insufficient over time. The challenge lies in continuously updating foundation models to accommodate novel information while retaining their original capabilities.  Leveraging the fact that foundation models have initial knowledge on various tasks and domains, we propose a novel approach that,  instead of updating all parameters equally,  localizes the updates to a sparse set of parameters relevant to the task being learned. We strike a  balance between efficiency and new task performance, while maintaining the transferability and generalizability of foundation models.
We extensively evaluate our method on foundational vision-language models
with a diverse spectrum of continual learning tasks. Our method achieves improvements on the accuracy of the newly learned tasks up to 7\% while preserving the pretraining knowledge with a negligible decrease of 0.9\% on a representative control set accuracy. Code is available here: \url{https://github.com/wx-zhang/spu}
\end{abstract}

\section{Introduction}
\begin{figure}
\centering
\includegraphics[width=\linewidth]{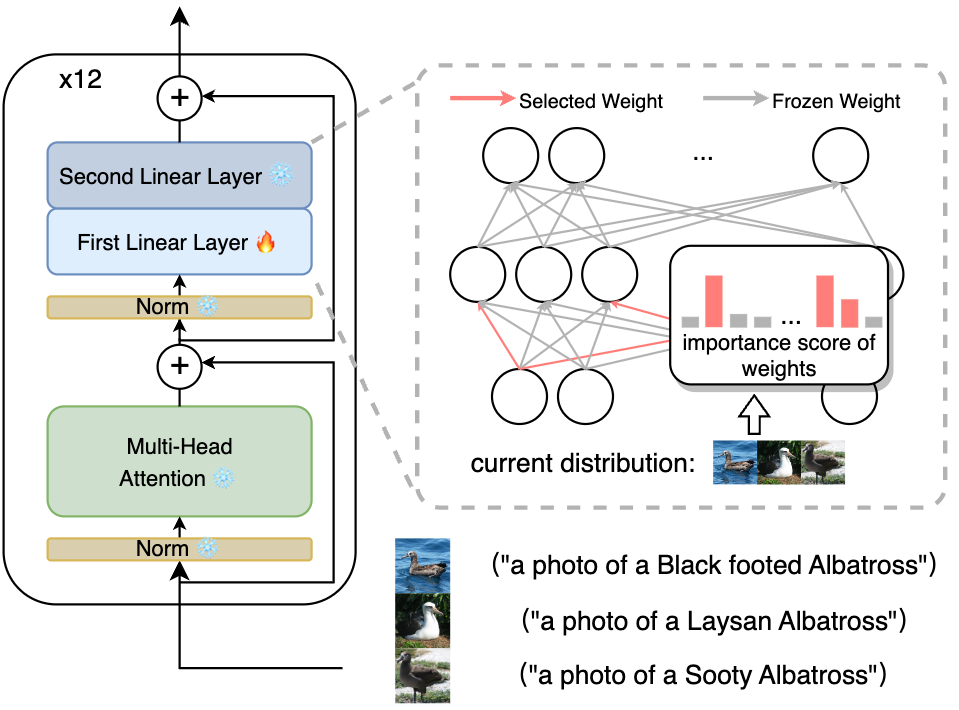}
\caption{We propose SPU algorithm. We first localize our update to the first layer of MLP blocks, and then select a sparse set of parameters specialized to the new task to update.}\label{fig:method}
\end{figure}
Recent machine learning models trained on a broad dataset have shown remarkable success in both natural language processing tasks~\citep{gpt4} and computer vision tasks~\citep{clip,flamingo}. These models can directly solve a wide range of tasks, such as recognizing common objects and answering common questions, thus are dubbed as foundation models~\citep{bommasani2021opportunities}. 
What is captured by these models covering various domains and tasks can be referred to as generic knowledge. 
Despite this, 
foundation models could still perform poorly on specific tasks.
For instance,~\citet{loraewc} found ChatGPT  limited in embodied tasks, while CLIP~\citep{clip} is shown struggling in recognizing fine-grained classes like cars from different brands. 
Therefore, it is crucial to integrate newly revealed data with pre-trained foundation models and expand their knowledge base. 
As one common solution, finetuning foundation models on new data would usually result in a good performance on the new task if done carefully. This will turn the foundation model into a specific model for a specific task, and would risk losing the existing capabilities of the model or the generic knowledge it has acquired through long phases of pre-training. The effect of deteriorating the model's previous knowledge upon new learning is a typical phenomenon of neural networks, referred to as catastrophic forgetting~\citep{mccloskey1989catastrophic}.

Continual learning research has been exploring the problem of  accumulating knowledge without forgetting~\cite{parisi2019continual} over the past years and has provided valuable techniques. However, most existing works
consider this process starting from a randomly initialized model~\citep{ebrahimi2019uncertainty,acl}.
Recently, with the success of large pre-trained models~\citep{steiner2021augreg, rw2019timm},
many works have considered continual learning starting from a pre-trained model~\citep{l2p,dualprompt}. 
Nevertheless, the emphasis lies mostly on the learning and forgetting behavior of the newly acquired knowledge, in the upcoming task sequence, often side-lining the pre-trained knowledge.  
Generic knowledge embedded in large models provides bases for strong performance in various domains and quick transfer to different tasks; when continuously finetuning a large pre-trained model on newly received tasks with no regard to preserving its pre-existing knowledge, we are losing the pre-training benefits and being left with merely a large model to deal with.

These prompt a crucial question: Can we effectively and continuously update foundation  models while retaining their generic knowledge? 
An example is accommodating a generic multi-modal model like CLIP~\citep{clip} to specific fine-grained concepts as various types of vehicles  while maintaining its generic recognition capabilities of common world concepts such as people, animals, and plants.

Towards this goal,
we seek to update  foundation vision-language models from a continual learning perspective while preserving their previously acquired generic knowledge. 
Starting from a large model pre-trained on vast sources of data, it is reasonable to assume that the model has some kind of basic or related knowledge on the new upcoming data. Thus, we hypothesize that there is an implicit modularity in the foundation model and design a method to locate which parameters are most relevant to the new upcoming data. 
Formally, we first identify specific model layers to be updated based on model analysis works~\citep{analysis-embedingspace,geva2020transformer}. Among the localized layers, we propose a mechanism to select parameters that are specialized for the task at hand. We opt for selecting parameters that small changes to their values would contribute to a greater improvement in the new task performance compared to   other parameters.
By doing so, we localize and update only a small number of the selected  parameters, while keeping a large portion of the model's parameters untouched. In this way, we not only provide an efficient method to finetune a large pre-trained model on newly arriving data but also preserve greatly the generalizability and transferability of the model.
Our strategy is to be executed whenever new data corresponding to a new set of classes, a new task or domain, is received.

To facilitate a comprehensive analysis of the generic knowledge deterioration, we focus on the classification tasks and formulate the knowledge base as the zero-shot classification ability on a diverse control set containing a wide range of classes. Our main objective is to demonstrate an improvement of a pre-trained model's performance on datasets where it initially exhibits suboptimal results, while preserving its original ability on a control set,  without revisiting it. 
We evaluate our method on six continual learning tasks and find that by updating merely 3\% of the parameters, our approach achieves performance on the new tasks superior to that achieved by methods that fully finetune the model, with almost no deterioration on the generic knowledge, only 0.97\% performance loss on the control set.  We further conduct comprehensive analyses to assess the impact of each component on generic knowledge forgetting. 

Our contribution can be concluded as 
1) We introduce the evaluation of generic knowledge forgetting in  continual learning, starting from foundation models. 2) To ensure the preservation of pre-trained knowledge, we propose an efficient method that localizes the learnable parameters,  selects specialized parameters for the new coming data, and performs sparse updates. 3) Through comprehensive evaluations on six datasets, we demonstrate that our algorithm significantly expands the pre-trained knowledge on new tasks while still preserving the generic knowledge. Additionally, we conduct in-depth analyses to understand the impact of each component on generic knowledge forgetting.

\section{Related Work}
\textbf{Foundation Models.}
pre-training techniques have played a crucial role in establishing the so-called foundation models, such as CLIP~\citep{clip}, Flamingo~\citep{flamingo}, BLIP-2~\citep{blip2}, PaLM-E~\citep{palme}, and GPT-4~\citep{gpt4}. These models are pre-trained on vast and diverse datasets, providing them with a broad knowledge base and exceptional generalization and transferability. Consequently, many of these models can be directly applied to various tasks in a zero-shot manner. Despite their strong abilities, evaluating these foundation models remains challenging~\citep{xu2023lvlm}, given that their strengths lie predominantly in a diverse domain of generalization.
While CLIP~\citep{clip}, an early vision-language model pre-trained on a large dataset of 400 million images and text samples, namely WebImageText, is an exception that exhibits impressive performance mainly on zero-shot classification tasks. This straightforward evaluation format allows us to thoroughly explore the changes in the model's knowledge base when implementing updates or modifications. By studying the impact of these changes on CLIP, we aim to gain a more in-depth understanding of the potential of updating the foundation models.

\textbf{Continual Learning.}
In the realm of continual learning, early methods \citep{er, ewc, agem, acl} train models from scratch for each specific sequence. Recent methods leverage the power of  pre-trained models to handle a new sequence of tasks.  Piggyback~\citep{piggyback}, as a pioneer, learns separate masks over a frozen pre-trained model for different tasks in the sequence. It requires storing the masks and access to task identification  to apply the mask during  inference, which is a limiting assumption.   Another line of work introduces additional parameters to acquire new knowledge~\citep{dualprompt, l2p, sprompt, smith2023coda}. Determining which set of newly added parameters to use during inference remains challenging. Additionally, the performance of such works is highly dependent on the capacity and flexibility of the added parameters, where some works  only get a marginal improvement over the pre-trained model~\citep{janson2022simple}.  Our work focuses on modifying the pre-trained models themselves, and shares some similarities with weight regularization methods~\cite{ewc,mas} where an importance or relevance score is estimated for the model's parameters. A clear distinction is that the parameter importance score is estimated \emph{after} learning a given task and used to \emph{prevent} changing those important parameters. Differently, our approach estimates the parameter's relevance score for a new task \emph{before} starting the learning process. Our selection is to identify which parameters to  \emph{update}. 
Finally,    the majority of these approaches focus on defying forgetting in the learned sequence, with no consideration for the forgetting of pre-trained knowledge. Further, they do not scale to preserving pre-trained knowledge,  as they either require access to the pre-training dataset~\citep{mas,ewc, er} or a duplicate storage of the pre-trained model~\citep{lwf,prd}. In contrast, we consider the accumulation of knowledge, including the pre-trained and newly acquired knowledge, without any task identification and extra storage of model weights.

\textbf{Finetuning with Knowledge Preservation.}
It is usually observed that when finetuning foundation models on new tasks, the generic knowledge and transferability are severely deteriorated.
Recently, some works~\citep{chen2022visualgpt, meng2022locating,ilharco2022editing,loraewc,khattak2023self,zheng2022preventing} started to tackle the issue of updating large pre-trained models while preserving their transferability and the generalizability. Among them,~\citet{meng2022locating, ilharco2022editing} proposes model editing algorithms, where the models are first analyzed to pick specific layers to edit, and then algebra-based or meta-learning based methods are applied to the weight of the local layer. Usually, a local set is utilized to preserve the background knowledge. 
While these methods have shown promise in incorporating specific concepts into the model, their impact on the generic knowledge remains uncertain, as discussed by~\citet{onoe2023can}. Additionally, most of these techniques are designed for specific models
for small-scale sample-wise edit of concrete mistakes and updates. Moreover,
they are centered around language models, where the input data has a stronger relationship to the concept being edited, leaving the vision models, where the input images can contain various of unrelated visual concepts, relatively unexplored. In contrast, we are interested in allowing continuous model updates on a set of new coming data samples, which can be scaled up to a larger number of concepts and a longer never-ending sequence. 

Additionally,~\citet{loraewc} proposed to finetune language models for embodied tasks while maintaining their generalization ability to handle unseen embodied tasks.  They suggested fine-tuning language models with LoRa~\citep{lora}, i.e., low rank updates, to ensure compute efficiency, while applying EWC regularization~\cite{ewc} to reduce forgetting of the pre-trained knowledge. On the multi-modal models end,~\citet{zheng2022preventing} considered to prevent zero-shot transfer degradation in the continual learning of CLIP by performing distillation on the pre-trained model weights. However, it requires access to a massive dataset to represent the pre-training distribution, which is not a trivial assumption and far from being computationally efficient. In this work, we aim to update foundation models, such as CLIP, continually to recognize additional concepts and preserve their transferability, while striving for efficiency.

 \section{Continual Learning From Pretrained Models} \label{sec:setting}

In Class Incremental Learning (CIL), we are given a dataset $D^t_\text{train} = \{ x_{k}, y_{k} \}_{k=1}^{N_t} \sim \mathcal{D}^t$ sampled from a task-specific distribution $\mathcal{D}^t$ for each task $t \in \{1, \ldots, T\}$ sequentially, where $X^t_\text{train} = \{ x_{k} \}_{k=1}^{N_t}$ is a set of images and $Y^t_\text{train} = \{ y_{k} \}_{k=1}^{N_t}$ is the set of the corresponding labels with $y_{k} \in Y^t_\text{train}$. Here $Y^t_\text{train}$ is the label space of task $t$.  Note that while we focus on image-based data, our method can be extended to any modality.
We are given a model parameterized by $\theta$  pre-trained on a vast pre-training dataset  $D_p \sim \mathcal{D}_p$ sampled from the pre-training distribution, which is inaccessible during the CIL procedure. During the learning of each task, the model parameters $\theta$ are to be optimized to minimize a loss function $\mathcal{L}$  on the current training set $D^t_\text{train}$. The loss function depends on the task at hand and the model deployed. For  CLIP model~\citep{clip} and image text pairs data, we deploy the same contrastive loss used for CLIP pre-training.  After the learning of each task, we evaluate our model on both the validation set of the seen distributions of the CIL sequence $D_\text{test}^{1:t}$, where $D_\text{test}^{t} \sim \mathcal{D}^t$, and a small control set $D_\text{control} \sim \mathcal{D}_p$ sampled from the pre-training distribution.

\section{\underline{S}elective \underline{P}arameter \underline{U}pdate (SPU)}
Most existing continual learning methods that start from randomly initialized models, which cannot provide prior to the task being learned, optimize all parameters equally. However, foundation models often have a reasonable initial performance on novel tasks, indicating some pre-existing knowledge relevant to these tasks. With the thriving for efficiency and the preservation of the generic knowledge, we suggest identifying 
a small set of parameters corresponding to tasks in hand and only updating them instead of modifying all the pre-trained model parameters. 
We now introduce how to localize the update to specific layers and how to identify a sparse set of specialized parameters to be optimized.

\label{sec:method}

\textbf{Localization.}
The objective of our work is to accumulate new knowledge without catastrophically forgetting the generic knowledge. To achieve this, we introduce a method that performs local changes restricted to specific layers in the pre-trained transformer backbones.
As shown in \cref{fig:method}, each layer of a transformer model is a transformer block, and  a transformer block contains a multi-head attention block and a two-layer MLP block. 

\begin{figure}
    \centering
    \includegraphics[width=0.94\linewidth]{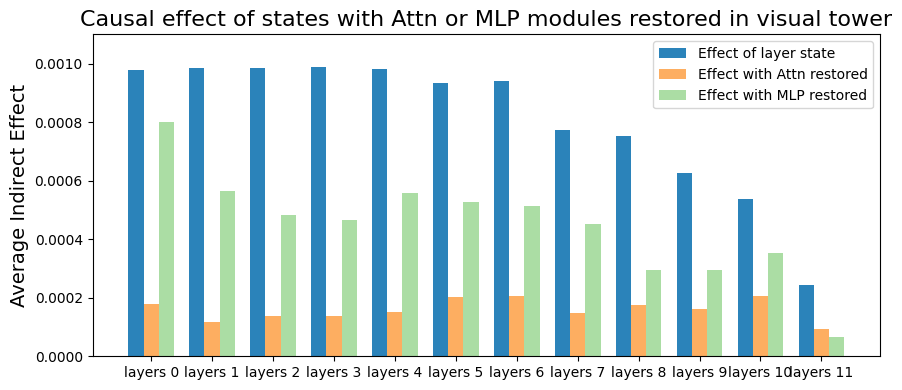}
    
    \includegraphics[width=0.94\linewidth]{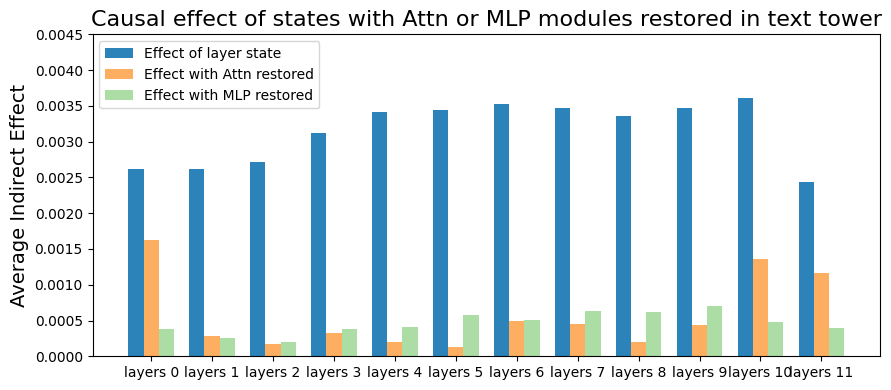}
    \caption{Casual tracking results of visual and text tower of CLIP. Changing MLP layers has a higher effect on the CLIP prediction results than changing Attention layers. }\vspace{-5mm}
    \label{fig:casual}
\end{figure}
\citet{meng2022locating} adopted casual tracking, widely adopted by later works~\citep{meng2022locating,hartvigsen2022aging,mend,meng2022mass},  to analyze the contribution of  attention layers and MLP layers to the output prediction. It performs comparisons by computing the average effects of restoring activation values at these locations over a corrupted input. More details of the causal tracking can be found in  \cref{appen:casualtracking}.
We follow the casual tracking analysis   and show, in \cref{fig:casual}, that the changes on the first MLP layer, that we localize the update to, have a larger effect on the model predictions than changes in attention layers.

\citet{geva2020transformer} further shows that MLP blocks emulate key-value neural memories, where the first layer of MLP acts as memory keys, operating as pattern detectors. Each individual key corresponds to a specific pattern seen in the input data. Whereas, the second layer learns the distribution over the detected patterns.
Our work aims to add, update, or refine current knowledge embedded in the model, and with the analogy to the key-value memories, we opt for refining the keys (corresponding to pattern detectors) to accommodate the new information. 
Empirically, we investigated whether we need to change the patterns' distributions represented by the second MLP layer and attention layer as well,  and it turned out that updating the first layer is sufficient and more effective, as we shall show in the  \cref{sec:ablation}. 

With the above in mind, we localize the model updates to the first layer of the MLP in each transformer block. With such localization, our candidate parameters to change can be reduced to  only around one third of the total parameters.

\textbf{Parameter Selection. }
Pre-trained foundational models have inherent knowledge, as evidenced by their capacity to execute diverse tasks without fine-tuning. Moreover, recent investigations~\citep{geva2020transformer,residual-empirial,bills2023language} have unveiled the correlation between the concepts and specific neurons' output in foundation language models. Therefore, we hypothesize that there exists modularity and specialization among specific neurons and their corresponding parameters in foundation models. Updating the most related neurons while keeping other neurons unchanged will not only facilitate the learning of new tasks but prevent the inference between different concepts in the new task sequence and between newly learned concepts and ones learned from pre-training.

Upon these, we propose to identify which parameters in the first MLP layer are specialized on the task at hand before training. As shown in \cref{fig:method}, the selection is associated with a scoring function, and we later minimize the new task loss by \emph{only} updating those selected parameters.

Formally, we receive the current task dataset $D^t$ representing a  task $t$ in a continual learning sequence and localize the updates to the first MLP layer $\theta^l$  
for each transformer block, where $l$ denotes the localized first layer  indexed over transformer blocks. We aim to define an element-wise scoring function $\mathcal{S}(\theta^{l}_{i,j}, D^t)$,
for each  parameter in a localized layer $\theta^{l}_{i,j}$;
 $i,j$ refers to the parameter connecting an input element $i$ (the $i$-th output entry of the attention layer) to the neuron $j$ in the first MLP layer.
We propose to select a subset of parameters $\theta_U^{l}\subseteq \theta^{l}$ that has the largest scores $\{\mathcal{S}(\theta^{l}_{i,j}, D^t)\}$, subject to $\frac{|\theta^{l}_U|}{|\theta^{l}|} =r$, where $|\cdot|$ is the parameter size and $r$ is the selection rate. 
This set is then expected to combine the most relevant parameters to the current task, represented by the dataset $D^t$. We select parameters regardless of their corresponding neurons and ablate the effect of selecting the entire parameters of identified neurons in \cref{appen:fullresults}. 
For clarity,  the presentation of the method is focused on $\theta^l$, and it can be generalized to a plural of selected layers covering all transformer blocks.

The idea of updating a sparse set of parameters is also adopted in related fields. We further comment on the relations and differences of these works in \cref{appen:relation}.

\textbf{Gradient-Based Scoring Function.} 
We aim to identify which parameters are more relevant to the new task at hand by this scoring function. We formulate this as finding parameters where small changes to their values could lead to a greater improvement in the task performance, with the loss function as a proxy.  When achieving this, we only make small changes to the model and thus can preserve the generic knowledge while improving the new task performance.  Specifically, we can approximate the change in the loss function $\mathcal{L}$ upon small changes $\delta$ in the parameters' values with
\begin{equation}\label{eq:1}
   \mathcal{L}(\theta^l+ \delta;x_k) - \mathcal{L}( \theta^l;x_k)\approx \sum_{i,j} g_{ij}(x_k)  \delta_{ij} = \frac{\partial (\mathcal{L}( \theta^l;x_k))}{\partial \theta^l_{ij}} \delta_{ij},
\end{equation}
where  $g_{ij}(x_k)$ is the gradient of the loss function regarding the parameter $ \theta^l_{ij}$  evaluated at the data point $x_k \in D^t$, and $\delta_{ij}$ is the local change in parameter space.
The above first-order approximation suggests that a fixed small changes made to parameters with the larger gradient magnitude $\parallel g_{ij}\parallel$ in the opposite direction of the gradient would incur a larger reduction in the loss function, and hence greater improvements with minor changes.

Following this, we define our scoring function as:
\begin{equation}\label{eq:2}
\mathcal{S}(\theta^l_{ij}, D^t)=\|\frac{1}{N'_t}\sum_{k=1}^{N'_t}   g_{ij}(x_k) \|,
\end{equation}
where $N'_t$ is the number of samples we use to compute the gradient.  $N'_t$  can be much smaller than the total number of samples in the dataset, $N_t$, as shown in the \cref{appen:fullresults}.

\textbf{Sparse update. }
Upon selecting the relevant parameters $\theta_U=\{\theta^l_U\}$, we freeze all other model parameters and learn the current dataset $D^t$ by  only optimizing  $\theta_U$.

Following the current practice in class incremental  learning methods,~\citep{prd,acl,l2p} we deploy a replay buffer to reduce the forgetting across the new tasks sequence. 
We  keep a replay buffer $\mathcal{M}$ of a fixed size, and sample batches from it of the same size as the batch from the current dataset at each optimization step. We update the replay buffer at the end of learning of each task by experience replay~\citep{er}.

Our final objective function at task $t$ can be written as %
\begin{equation}
    \min_{\theta_U} \mathcal{L}(\theta; D^t_\text{train}) + \mathcal{L}(\theta; \mathcal{M})
\end{equation}
where $\mathcal{L}(\theta; D)$ is the  loss computed on the set $D$.  

\textbf{Algorithm applicability.} Our algorithm involves three key steps: localizing update layers, selecting relevant parameters, and training on the new task with sparse updates. It is important to note that while we primarily delve into the localization within the transformer architecture, the concept of selectively updating certain layers while keeping others frozen to achieve efficiency and comparable performance is  not confined to this architecture alone.~\citep{santurkar2021editing,osaka}. Should the need arises to extend our approach to different architectures, the first step of our methodology can be readily adapted.  Furthermore, the processes of parameter selection and sparse updates remain architecture-agnostic, making them versatile across various model structures. 

\begin{table*}[]
\centering 
\resizebox{\linewidth}{!}{
\begin{tabular}{l|ccc|ccc|ccc|ccc|ccc|ccc|  ccc}
\toprule
\multicolumn{1}{l}{} & \multicolumn{3}{c}{Aircraft} & \multicolumn{3}{c}{Birdsnap} & \multicolumn{3}{c}{Cars} & \multicolumn{3}{c}{CIFAR100} & \multicolumn{3}{c}{CUB} & \multicolumn{3}{c}{GTSRB} & \multicolumn{3}{c}{Average} \\\midrule
\multicolumn{1}{l}{} & Acc. & F. & C. & Acc. & F. & C. & Acc. & F. & C. & Acc. & F. & C. & Acc. & F. & C. & Acc. & F. & C. & Acc. In. & Avg. F. & C. Drop \\\midrule
Frozen~\citep{clip} & 24.45 & - & 63.55 & 43.20 & - & 63.55 & 64.63 & - & 63.55 & 68.25 & - & 63.55 & 55.13 & - & 63.55 & 43.38 & - & 63.55 & 0.0 & - & 0.0 \\\midrule
FLYP~\citep{goyal2023finetune} & 18.63 & 39.93 & 41.04 & 44.06 & 23.43 & 51.06 & 51.64 & 25.65 & 52.25 & 46.26 & 37.78 & 26.53 & 45.74 & 26.62 & 44.30 & 21.76 & 55.48 & 1.59 & -11.82 & 34.81 & 27.42 \\
\,+ MAS~\citep{mas} & 33.69 & 27.50 & 61.09 & 47.42 & 17.12 & 60.05 & 69.43 & 9.18 & 61.17 & 63.88 & 21.16 & 49.35 & 61.72 & 12.05 & 57.35 & 42.04 & 25.38 & 42.06 & 3.19 & 18.73 & 8.37 \\
\, + ER~\citep{er} & 41.42 & 31.48 & 50.41 & \textbf{56.22} & 21.63 & 56.72 & 69.08 & 16.42 & 58.07 & 82.86 & 3.41 & 42.10 & 64.07 & 17.72 & 51.30 & \textbf{96.28} & -7.40 & 17.34 & 18.48 & 13.88 & 17.56 \\
\, + ER  + LwF~\citep{lwf} & 36.08 & 18.12 & 63.06 & 50.23 & 10.20 & 62.08 & 72.56 & 4.04 & 62.59 & 74.32 & 8.16 & 55.71 & 65.11 & 5.90 & 62.05 & 53.56 & 11.86 & 57.99 & 8.80 & 9.71 & 2.97 \\
\, + ER + PRD~\citep{prd} & 37.11 & 17.35 & 63.38 & 51.34 & 9.45 & 62.85 & 74.08 & 3.75 & 62.96 & 79.66 & 3.10 & 59.01 & 65.92 & 6.55 & 62.09 & 63.00 & 12.44 & 61.04 & 12.01 & 8.77 & 1.66 \\\midrule
 LoRA-EWC~\citep{loraewc}& 30.36& 12.23& 62.82& 45.91& 12.12& 62.53& 66.11& 3.89& 61.39& 67.35& 15.28& 55.27& 58.72& 4.92& 61.27& 46.14& 13.23& 61.70& 2.59& 10.28&2.72\\
\, + ER (r=8)& 33.12& 12.14& 62.99& 50.28& \textbf{8.70}& 62.74& 70.51& 0.88& 62.49& 81.27& -0.70& 59.90& 62.36& 2.99& 62.80& 89.87& -7.17& 61.62& 14.73& 2.81&1.46\\
 \, + ER (r=96)& 33.75& 11.75& 62.91& 50.52& 8.96& 63.00& 71.17& \textbf{0.46}& 62.39& 82.10& \textbf{-1.59}& 59.91& 62.31& \textbf{2.81}& 62.72& 90.01& -7.46& 61.66& 15.14& \textbf{2.49}&1.45\\
\midrule
L2P~\citep{l2p} & 32.20 & 21.73 & 43.43 & 24.37 & 36.17 & 44.63 & 67.04 & 11.22 & 42.53 & 67.71 & 18.81 & 39.61  & 64.04 & 6.82 & 45.51 & 75.45 & 2.68 & 34.05 & 5.29& 16.24&  21.92\\
DualPrompt~\citep{dualprompt} & 26.61& 17.20& 56.31& 36.34& 30.23& 46.43& 63.30& 18.67&55.76& 61.72& 19.87& 42.37& 64.38& 12.94& 55.63& 69.65& 8.43& 40.37& 3.83& 17.89&  14.07\\
SLCA~\citep{slca} & 29.40& \textbf{11.45}& 63.49& 43.18& 9.28& \textbf{63.33}& 62.65& 4.42& 63.29& 70.03& 0.19& 60.23& 53.87& 7.75& 63.31& 46.01& 0.83& 62.76& 1.02& 5.65&0.81\\
ZSCL~\citep{zheng2022preventing} & 30.96 & 15.65 & \textbf{65.53} & 49.85 & 13.28 & 63.13 & 67.79 & 8.27 & 62.90 & 80.50 & 1.05 & \textbf{61.90} & 61.09 & 7.69 & 62.78 & 62.92 & 13.54 & \textbf{62.92} & 9.01 & 9.91 & \textbf{0.36} \\
SparseCL~\citep{wang2022sparcl}& 31.95& 19.77& 63.31& 45.11& 16.78& 61.50& 71.57& 5.38& 62.82& 69.35& 15.23& 57.39& 62.50& 9.66& 62.43& 48.99& 24.91& 61.03& 5.07& 15.29& 2.14\\
SPG~\citep{konishi2023parameter}& 39.15& 21.42& 63.62& 49.25& 14.88& 62.55& 73.09&  5.94& 63.30& 69.79& 14.99& 59.40& 65.43& 8.18& 62.43& 54.36& 17.73& 61.74& 8.67& 13.86& 1.38 \\
\midrule
\textbf{SPU} - Ours & \textbf{44.43} & 14.42 & 63.48 & 55.35 & 12.78 & 61.94 & \textbf{77.51} & 3.26 & \textbf{63.42} & \textbf{83.99} & -0.39 & 61.38 & \textbf{71.51} & 4.84 & \textbf{62.87} & 94.25 & \textbf{-7.87} & 62.55 & \textbf{21.34} & 4.51 & \textbf{0.94}\\
\bottomrule
\end{tabular}}
\caption{  Average Accuracy (Acc.), Forgetting (F.), and control set Accuracy (C.) of our method \ours and baselines on  6 CIL sequences, demonstrating our superior knowledge accumulation and preservation. We highlight  parameter efficiency via parameters size and learnable parameters rate, and data efficiency via data use. }
\label{tab:cilresults}
\end{table*}

\section{Experiments}
We evaluate our proposed 
framework  on various datasets compared to different methods and baselines in \cref{sec:cil-exp}, and analyze different components of our method and ablate our design choices in \cref{sec:ablation}.
We provide further ablations on defying generic knowledge loss in   the \cref{appen:fullresults}.

\subsection{Setup}
\myparagraph{Backbone.} We apply SPU to vision-language classification tasks, given the   relatively robust measurement of the knowledge base in such tasks.  We choose the pre-trained CLIP-ViT/B-16~\citep{clip} as our backbone.

\myparagraph{Datasets.} We evaluate the performance of our algorithms on a total of six datasets— four fine-grained datasets (Birdsnap~\citep{birdsnap}, CUB-200-2011~\citep{cub},  FGVC-Aircraft~\citep{aircraft}, Stanford Cars~\citep{cars}), one coarse dataset (CIFAR100~\cite{cifar100}), and one out-of-distribution dataset (GTSRB~\citep{gtsrb}). These datasets are chosen primarily based on their initially low zero-shot performance with CLIP pre-trained models.
To form the continual learning sequences, we split each dataset into 10 subsets with disjoint  classes composing 10 tasks. For methods that leverage a replay buffer, we use a buffer size of around 4\% of the dataset size. Ablation study of buffer size is shown in \cref{sec:ablation}. For more comprehensive information, please refer to the \cref{appen:implementation}.

\myparagraph{Baselines.}  We conduct a comprehensive comparison of our method against various baselines. 
Firstly, we evaluate our approach against the best fine-tuning method of CLIP, FLYP~\citep{goyal2023finetune}. We further integrate with FLYP classical continual learning components  to evaluate their performance on the CLIP backbone, including ER~\citep{er}, weight regularization method, MAS~\citep{mas}, and functional regularization methods LwF~\citep{lwf} and PRD~\citep{prd}.  We combine these functional regularization methods with a replay buffer.
We further consider the latest pre-trained model based continual learning techniques.
  L2P~\citep{l2p}, DualPrompt~\citep{dualprompt}, and SLCA~\citep{slca}.
  Finally, we  compare to two recent methods that target knowledge retention of foundation models. 
   ZSCL~\citep{zheng2022preventing} designed for CLIP~\cite{clip}
and   LoRA-EWC~\citep{loraewc} which combines LoRA~\cite{lora} and EWC~\cite{ewc} to finetune an LLM, here we adapt it to CLIP. 
Results, evaluation with  ImageNet pretrained backbones of these methods,  and discussion are in the \suppl.

\myparagraph{Evaluation Metrics.}  We measure the Acc. at the end of the class-incremental process, as well as the forgetting rate following prior arts~\citep{er,agem}. Additionally, 
we aim to understand how the knowledge base shifts as we continually update the pre-trained models. To achieve this, and similar to~\citep{ilharco2022editing}, we evaluate a continually trained model $M$ on a diverse dataset representing  generic knowledge, \ie, the validation set of ImageNet~\citep{imagenet}, which acts as a control set (C.). We report the zero-shot classification accuracy on (C.),  and compare it with that from the frozen pre-trained models. 

To provide a comprehensive view of model  performance across all $N_D$ datasets $\{D_i\}_{i=1}^{N_D}$, we denoted the model parameters trained after $D_i$ as $M_i$, and frozen model performance on $D_i$ as $ M_{\text{f}_i}$.  We present the increment of Average Accuracy (Acc. In.) across these datasets as
\begin{equation}\footnotesize
    \text{Acc. In.}(M) = \frac{1}{N_D} \sum_{i=1}^{N_D} \text{Acc.}(M_i) - \text{Acc.}( M_{\text{f}_i}),
\end{equation}
the average forgetting rate (Avg. F.) 
\begin{equation}\footnotesize
    \text{Avg. F}(M) = \frac{1}{N_D} \sum_{i=1}^{N_D} \text{F. } ( M_i ),
\end{equation}
and the average drop of control set accuracy (C. Drop).
\begin{equation}\footnotesize
    \text{C. Drop}(M) = \frac{1}{N_D} \sum_{i=1}^{N_D}\text{C.}(M_\text{f}) - \text{C.}(M_i),
\end{equation}

\myparagraph{Implementation Details.}
We follow~\citep{goyal2023finetune} to both perform selection and sparse update on the visual tower and text tower of the CLIP model, and use contrastive loss as our loss function. Within our algorithm, we use a selection rate of 10\%, which optimally balances learning and forgetting. We perform an ablation study on the selection rate in \cref{sec:ablation}. More implementation details is in~\cref{appen:implementation}.

\subsection{Results}\label{sec:cil-exp}

We present the comparison between our approach and other methods in \cref{tab:cilresults}. 
In the subsequent sections, we delve into our observations from the dual lenses of learning and forgetting.

\textbf{Comparison with other methods.}
We view accumulating novel knowledge as prioritized, at the same time also pay attention to knowledge retention.
Regrading the accuracy of newly learned knowledge (Acc.), we achieve state-of-the-art results in four out of six datasets, \ie, Aircraft, Cars, CIFAR100, and CUB,  and comparable results in Birdsnap and GTSRB,
with a notable average margin of 2.86\%
over the existing continual learning methods. We analyze how our scoring function contribute to the achievement in \cref{sec:ablation}.

Regarding the knowledge retention, our approach achieves  control set accuracy drop (C. Drop) of 0.94\% which is the least drop among  methods  with no external data access, and is comparable to that of ZSCL, which requires access to the additional Conceptual Caption~\citep{changpinyo2021cc12m} dataset for knowledge retention. This brings efficiency concerns, which we will elaborate later in \cref{sec:efficient}.
Meanwhile, ZSCL preserves  the generic knowledge at the expense of the new tasks increment average accuracy which is ~12\% lower than ours.

The superior results in new task accuracy and control set accuracy demonstrates that \ours can effectively extends the knowledge base during continual learning. 
 
Among the continual learning methods, FLYP+ER stands as the only comparable contender in terms of average accuracy of new task. This mainly benefits from the balanced loss terms on  buffer data and current task data.  However,  it exhibits a significant drawback in  forgetting, averaging at 13.88\% in the forgetting of the current dataset, and a notable decrease of 17.56\% in average control set accuracy. 

Distillation-based methods like FLYP + ER + LwF/PRD and ZSCL  generally perform good at preserving the pre-trained knowledge,  all  displaying control set accuracy drop of less than  3\%. 
However, their  flexibility in learning the new tasks, as indicated by their average accuracy, remains  limited,  reflecting a discernible gap of over 8\% when compared to our method. While SLCA achieves the second best results of 0.81\% in preserving the pre-trained knowledge,  it almost cannot improve the new task when compared to \ours.

LoRA based methods exhibit extraordinary performance in eliminating forgetting. In the forgetting of the new tasks, LoRA-EWC combined with ER can achieve only 2.49\% of forgetting. LoRA-EWC has only  1\% - 3\% control set accuracy drop depending on the rank choice and buffer choice. However, this is a larger drop than our marginal 0.94\% drop in control set accuracy. In spite of their knowledge retention ability being slightly worse than ours,  their average increment accuracy on new tasks is lower than ours, with at least a margin of 6.2\%. 

\begin{table*}[]
\centering
\resizebox{\textwidth}{!}{
\begin{tabular}{l|ccc|ccc|ccc|ccc|ccc|ccc|  ccc}\toprule
 & \multicolumn{3}{c}{Aircraft} & \multicolumn{3}{c}{Birdsnap} & \multicolumn{3}{c}{Cars} & \multicolumn{3}{c}{CIFAR100} & \multicolumn{3}{c}{CUB} & \multicolumn{3}{c}{GTSRB} & \multicolumn{3}{c}{Average} \\\midrule 
  & Acc. & F. & C. & Acc. & F. & C. & Acc. & F. & C. & Acc. & C. & H. & Acc. & C. & H. & Acc. & F. & C. & Acc. In. & Avg. F. & D. Drop \\\midrule
 Attention layers & 41.34 & 14.68 & \textbf{64.02} & 55.22 & \textbf{11.73} & \textbf{62.53} & 76.35 & 3.61 & \textbf{63.73} & 84.00 & -0.35 & \textbf{62.49} & 70.99 & \textbf{4.03} & \textbf{63.39} & 92.41 & -8.23 & \textbf{63.20} & 20.21 & \textbf{4.24} & \textbf{0.32}\\
 \rowcolor{lightgray}
First MLP layers & \textbf{44.43} & 14.42 & 63.48 & \textbf{55.35} & 12.78 & 61.94 & \textbf{77.51} & \textbf{3.26} & 63.42 & 83.99 & -0.39 & 61.38 & \textbf{71.51} & 4.84 & 62.87 & \textbf{94.25} & -7.87 & 62.55 & \textbf{21.34} & 4.51 & 0.94 \\
Second MLP layers & 43.32 & \textbf{14.02} & 63.24 & 54.98 & 12.25 & 61.17 & 76.91 & 3.24 & 62.77 & 83.59 & \textbf{-0.42} & 59.57 & 70.00 & 5.40 & 62.19 & 93.32 & \textbf{-8.38} & 61.24 & 20.51 & 4.35 & 1.85 \\
Both MLP layers & 44.21 & 14.78 & 63.32 & 55.10 & 13.46 & 61.32 & 77.25 & 3.79 & 63.13 & \textbf{84.15} & -0.31 & 60.47 & 71.23 & 5.42 & 62.34 & 94.18 & -7.85 & 61.65 & 21.18 & 4.88 & 1.51 \\\bottomrule
\end{tabular}}
\caption{ Compared to various choice of selected layers, our choice (in gray background) achieves the best performance in new task accuracy (Acc.) while has comparable results in control set accuracy (C.) }
    \label{tab:updatelater}
\end{table*}

\begin{table*}[]
\centering
\resizebox{\textwidth}{!}{
\begin{tabular}{l|ccc|ccc|ccc|ccc|ccc|ccc|  ccc}\toprule
 & \multicolumn{3}{c}{Aircraft} & \multicolumn{3}{c}{Birdsnap} & \multicolumn{3}{c}{Cars} & \multicolumn{3}{c}{CIFAR100} & \multicolumn{3}{c}{CUB} & \multicolumn{3}{c}{GTSRB} & \multicolumn{3}{c}{Average} \\\midrule
 & Acc. & F. & H. & Acc. & F. & H. & Acc. & F. & H. & Acc. & F. & H. & Acc. & F. & H. & Acc. & F. & H. & Acc. In. & Avg. F. & C. Drop \\\midrule
Random & 38.34 & \textbf{11.19} & \textbf{63.83} & 54.74 & \textbf{8.92} & \textbf{63.64} & 74.62 & \textbf{2.91} & \textbf{63.64} & 83.84 & \textbf{-2.17} & \textbf{62.88} & 67.36 & \textbf{3.59} & \textbf{63.77} & 86.51 & -6.06 & \textbf{63.30} & 17.73 & \textbf{3.06} & \textbf{0.04}\\
\rowcolor{lightgray}
SPU & \textbf{44.43} & 14.42 & 63.48 & \textbf{55.35} & 12.78 & 61.94 & \textbf{77.51} & 3.26 & 63.42 & 83.99 & -0.39 & 61.38 & \textbf{71.51} & 4.84 & 62.87 & \textbf{94.25} & \textbf{-7.87} & 62.55 & \textbf{21.34} & 4.51 & 0.94 \\
piggyback~\citep{piggyback} & 43.68 & 14.86 & 63.66 & 53.81 & 13.97 & 61.91 & 76.58 & 3.94 & 63.45 & 83.93 & -0.70 & 61.45 & 70.97 & 5.00 & 62.98 & 93.02 & -7.83 & 62.30 & 20.49 & 4.87 & 0.92 \\
Mask & 43.95 & 14.80 & 63.58 & 54.23 & 13.10 & 62.23 & 76.92 & 3.56 & 63.43 & \textbf{84.30} & -1.07 & 62.08 & 71.11 & 4.78 & 62.98 & 92.41 & -7.33 & 62.63 & 20.65 & 4.64 & 0.73 \\
\bottomrule
\end{tabular}}
\caption{Compared to random selection, our superior performance (in gray background) implies the selected weights  specialized to the task. Compared to training-based scoring functions, our training-free function performs better in  new task accuracy and control set accuracy.  }
    \label{tab:strategy}
\end{table*}

\textbf{Fine-grained Datasets.}
The diverse characteristics of various datasets also lead to distinct behaviors. Across fine-grained datasets like Aircraft, Cars, and CUB, we achieve SOTA average accuracy, outperforming the baselines by around 3\%, while demonstrating minimal degradation in control set accuracy of less than 1\%.   

\textbf{Out of Distribution Dataset.} We view GSTRB 
as out of distribution for CLIP pretraining, as it is the only considered dataset where the  zero shot performance of CLIP is significantly lower than the performance of a linear classifier trained on ResNet50 features~\cite{clip}. Its extremely detailed class descriptions (such as ``blue circle with white forward arrow mandatory'') make the deep semantic understanding of images, such as the exact meaning of the signs, less important.  
In these experiments,  GSTRB proves  an outlier for SOTA CIL methods with significantly low Acc., while  our method proves robust. %
FLYP+ER achieves an average accuracy of 96.28\% in GTSRB, but at the expense of a 17.34\% control set accuracy, equating to around 60\% accuracy loss, indicating  a large decay in the generic knowledge after learning such out of distribution datasets. 
In contrast, our proposed method achieves competitive  accuracy, concurrently delivering small control set loss of around 1\%, signifying minimal loss of generic knowledge.

\textbf{Coarse Dataset.}
In contrast, in the case of the coarser CIFAR100 dataset, we still achieve an impressive SOTA learning accuracy of 83.99\%, albeit with a marginal trade-off of approximately 2\% in control set accuracy.  Even with this reduction, \ours stands out as significant compared to most other continual learning techniques that experience losses of generic knowledge ranging from 4\% to 21\%. This phenomenon can be attributed to that CIFAR100 encapsulates a degree of generic knowledge, possibly causing interference in the information on control sets like ImageNet.

\subsection{Ablation Study}
\label{sec:ablation}

In this section, we perform  ablation studies on the individual components comprising our algorithm  to validate the rationale behind our design of these components.
Refer to the \cref{appen:fullresults} for more details and full results.

\myparagraph{Which layer to update?} We compare localizing the update to  the first MLP layer parameters (our choice) to that of the second MLP layers and  both MLP layers together. We also consider the choice of Attention layers.
In the experiment of Attention layers and second MLP layers, we updated 10\% of parameters as what we do in our choice. In the experiment of updating both MLP layers, we updated 5\% parameters of each layer to match the selection rate.  
 Results reported in \cref{tab:updatelater}. 
Updating the Attention layers helps to migrate the forgetting better, which is consistent with the LoRA-EWC performance. However, it has obvious worse performances on the new tasks accuracy in Aircraft, Cars, CUB, and GTSRB. 
Updating parameters from the second layer  suffers double the generic knowledge loss compared to that of the first layer parameters.  Updating parameters in both layers is also worse in both forgetting and control set accuracy than that of the first layer only. We conclude that localizing the updates to  selected parameters of the first layer only is sufficient to achieve the best trade-offs.

\myparagraph{Do the selected weights represent the task?}
We validate whether the selected parameters can represent the  task at hand in \cref{tab:strategy} by comparing our scoring function with a random selection. The results indicate that  with sparse update, we can preserve the knowledge learned from pre-training. However, the Avg. In.   of the random select baseline, 17.73\%, is worse than the Avg. In. of FLYP+ER, 18.48\%. This suggests that  with only sparse update we may miss some important representations to the new task  in the parameter space. However, with our scoring function, we do not only improve over random select, but over full finetune (FLYP+ER) in continual learning by mitigating forgetting. This implies that our selected parameters are specialized in the current task concepts, thus changing them will cause the least interference with other tasks. In \cref{appen:selection}, we further visualize that the selected parameters can well represent the task, and we will select diffrent parameters for different tasks. 

There are also existing methods, \eg Piggyback~\citep{piggyback}, that train a mask for parameter selection. These methods require an additional phase of training; thus \ours is more computationally  efficient. Furthermore, in \cref{tab:strategy}, we compare \ours with Piggyback and a learnable variant of our scoring function, denoted as Mask. Details of the implementation is in \cref{sec:learnablescore}. Comparing with these two methods, our gradient-based scoring function is better in both new task learning (Acc.) and in knowledge preservation (C.).

\begin{table}[h]\centering\small
    \begin{tabular}{cccc}\toprule
Selection Rate & Acc. In. & Avg. F. & C. Drop \\ \midrule
 0.01 & 17.70 & \textbf{3.10} & 1.11 \\
 \rowcolor{lightgray}
 0.10 & 21.34 & 4.51 & \textbf{0.94} \\
  0.50 & \textbf{21.73}& 7.76 & 0.95 \\\bottomrule
\end{tabular}
\caption{Ablation on selection rate of SPU. Our approach achieves the best trade-off when selecting 10\% weights.}
        \label{tab:selectionrate}
\end{table}
\myparagraph{Selection rate.} 
\cref{tab:selectionrate} illustrates the variants of our method under varying selection rates applied to the first layer of MLP blocks. Across all selection rates, \ours demonstrates competitive average accuracy, forgetting, and control set accuracy when compared with other baselines in \cref{tab:cilresults}. Even with a 0.5 selection rate, the learnable parameters comprise only 30\% of the total parameters.  We note that as the selection rate  increases, there is a marginal enhancement in learning performance, but accompanied by a compromise in forgetting. For instance, raising from 0.1 to  0.5 selection rate, the Average Accuracy improves around 0.5\% but the forgetting also raises around 3\%.  Therefore, we opt for a selection rate of 0.1, which gives the best trade-off between the accumulation of the new knowledge and the preservation of the pre-trained knowledge.

\begin{table}[h]\centering
    \resizebox{\linewidth}{!}{
    \begin{tabular}{c|ccc|ccc}\toprule
 \multirow{2}{*}{\makecell{Buffer Size   \\ / Total Size}} & \multicolumn{3}{c|}{FLYP+ER}& \multicolumn{3}{c}{SPU}\\\cmidrule{2-7}
 & ACC. In& Avg. F. &C. Drop & ACC. In& Avg. F. &C. Drop \\ \midrule
   1\% & 8.97 & 22.27 & 19.18 & {16.18} & 10.28 &1.00  \\
 2\% &13.24 & 19.35 & 18.24  & 18.63 & 8.14 &0.96  \\
  \rowcolor{lightgray}
 4\% &{18.48} & {13.88} & {17.56}  & \textbf{21.34} & \textbf{4.51} &\textbf{0.94 } \\\bottomrule
\end{tabular}}
\caption{Ablation on buffer size and comparison to FLYP+ER. Our approach has lower performance drop  and small forgetting when the buffer size decreases}
    \label{tab:buffersize}
    \vspace{-3mm}
\end{table}
\myparagraph{Buffer size.} 
In \cref{tab:cilresults}, we present the outcomes of \ours using a buffer size equivalent to 4\% of the total dataset size.  \cref{tab:buffersize} shows our  performance over an array of buffer sizes, ranging from 1\% to 4\% of the total dataset size, compared with ER. Evidently, our algorithm excels in preserving pre-training knowledge across all buffer sizes, all with less than 1\% drop in control set accuracy. As we decrease the buffer size, FLYP+ER encounters substantial influence; our method with 1\% buffer size  doubles 
 Avg. Acc. improvement of FLYP+ER with 1\% buffer and suffers 50\% less  forgetting with merely 1\% control set accuracy loss.

\subsection{Efficiency}\label{sec:efficient}

\begin{table} 
    \centering \footnotesize 
    \begin{tabular}{lccc}\toprule
         Method& \makecell{Full \\  Parameter}&\makecell{Trainable \\  Parameter  }& \makecell{Extra Data \\ Source} \\\midrule
         FLYP &  149.5M &149.5M (100\%) & -\\
         LoRA-EWC (r=96) & 154M  &5.90M (3.79\%)&  CC12m\\
 ZSCL&299M &149.5M (50\%)& CC12m\\
         SPU (ours)&   149.5M&4.72M (3.15\%)&-\\\bottomrule
    \end{tabular}
    \caption{Parameter efficiency and data efficiency of various CL algorithms. Our approach is parameter and data efficient in updating a small portion of parameters with no added parameters and no requirement of extra data source. }
    \label{tab:efficiency}
\end{table}
We consider efficiency from two perspectives, parameter efficiency and data efficiency, as shown in \cref{tab:efficiency}.  For parameter efficiency, we follow~\citep{lora,he2021towards,jia2022visual} to report the full parameter size and trainable parameter size. 
While most of the current methods necessitate a complete parameter update, \ours only requires an update of a sparse subset of parameters, which only consists of 2.7\% of the total model's parameters. Besides this, we neither   require adding extra parameters to the model as in LoRA-EWC and L2P, nor storing the frozen pre-trained model as in ZSCL. Using  the pre-trained model consumes extra GPU memory during the training. Adding extra model parameters consumes extra GPU memory during the training, and disk memory when saving the model. This influence may be ignorable in a limited number of tasks. However,  continual learning expects an ever-going algorithm. Then the storage problem becomes profound, together with the added model components (prompt, adapter, and so on) choosing problem, as in L2P. 
 
For data efficiency, our algorithm does not require extra data source, making it light to deploy on various applications without loading huge datasets. LoRA-EWC and ZSCL are the only two methods achieving similar control set accuracy to \ours. However, LoRA-EWC takes Conceptional Caption 12M (CC12M)~\citep{changpinyo2021cc12m} to compute the Fisher information of pre-trained task, and ZSCL uses CC12M for distillation. 

We further perform an ablation study on the number of samples $N_t'$ used to approximate the scoring function. Results show that our method can still have good performance even when using only one batch of samples for the approximation. This implies that the computation of the scoring function  is also efficient which does not require a full pass of the data prior to the training, and can be done transparently with the first received batch.  Details are  in \cref{appen:efficiency}.

\section{Discussion}

With the rise of advanced foundation models pretrained on vast datasets, we propose a method that preserves pre-learned information in continual learning. We base on the fact that foundation models already have initial knowledge for the task in hand, and identify specific model layers and  parameters corresponding to this knowledge for sparse updates. As such, we perform small update for the model to cope with the new knowledge while preserving the previously acquired generic knowledge. We evaluate our method extensively and show superior performance.  However, our current method operates unidirectional, and future research could explore knowledge accumulation across diverse domains. Additionally, expanding our focus from discriminative to generative tasks would enhance the applicability of our techniques.

{
    \small
    \bibliographystyle{ieeenat_fullname}
    \bibliography{reference}
}
~\clearpage
\onecolumn
\appendix

\section{Casual Tracking for Localization} \label{appen:casualtracking}
In this paper, the  weight selection is localized to the first MLP layer within transformer blocks.  We discussed such localization in prior model editing and probing works in Section 4. We further perform casual tracking to  validate the localization. 

~\citet{vig2020investigating} quantifies the contribution of intermediate variables in causal graphs for causal mediation analysis. Based on this, \citet{meng2022locating} proposed casual tracking for identifying neuron activations that are decisive in a language model’s factual predictions. Casual tracking  identifies specific locations that  contribute to the input's recognition by computing the average effects of restoring activations at these locations on a corrupted input. We adapt the casual tracking to CLIP models and formulate the computation of average indirect effect (AIE) in the following. 

In CLIP model with ViT backbone, we freeze one tower of visual or text and perform casual tracking on the other. Take the casual tracking on image tower for example, we do the next three runs:
\begin{itemize}
    \item \textbf{Clean run. } We pass an image-text pair  into the model and store activations of the visual tower $\{a_i^\ell | i \in [1,T], \ell \in [1,L]\}$, and get the similarity score $S$. Here $T$ is the number of tokens, and $L$ is the number of layers.
    \item \textbf{Corrupted run. } We then pass the image into visual tower by adding noises to the image embeddings of patches related to the corresponding text and get a corrupted visual output feature. We compute the similarity score $S_c$ between this feature and the clean text features. 
    \item \textbf{Corrupted-with-restoration run. } Finally, we follow the corrupted run to add noises on image embeddings, and replace the activation of layer $\ell$ token $i$ with the clean activation $a_i^\ell$, and get corrupted-with-restoration visual output features. We compute the similarity score ${S_r}_i^\ell$ between these features and the clean text features.  
\end{itemize}
The average indirect effect (AIE) is computed by the average difference between the similarity scores of corrupted run and corrupted-with-restoration runs, \ie
\begin{equation}
    \text{AIE}^\ell = \frac{1}{T} \sum_{i \in [1,T]} \frac{\big| {S_r}_i^\ell - S_c \big |}{S}.
\end{equation}
Here $\big| {S_r}_i^\ell - S_c \big |$ measures the change of similarity scores when we restore one single state, \ie, activation, back to the clean activation. In CLIP model, we observe that this restoration often does not lead to positive effect to the similarity score; thus we compute the absolute change here. As we need to aggregate AIE over multiple image-text pairs, we normalize the change of similarity by the score from the clean run. In casual tracking of text tower, we freeze the CLIP visual tower and apply the same procedure on the text tower.

Intuitively, higher $\text{AIE}^\ell$ means activations or states of layer $\ell$ are more important to the final classification. We further compute AIE over MLP layers $\text{AIE}^\ell_\text{mlp}$ or Attention layers $\text{AIE}^\ell_\text{attn}$ by restoring activation values outputted from MLP or Attention layers among all the transformer blocks. 

In practice, we perform casual tracking on the validation set of COCO~\citep{lin2014microsoft} since it provides detailed information on objects in images. We decide the object-related image patch by the bounding box information. We use the prompt \texttt{a photo of \{class name\}} as text input, and the image-related tokens are those that represent the class name.   The casual tracking results are in \cref{fig:casual}. Here we show the effect of restoring states (activations) after full layer in blue, the effect of restoring states after Attention layers in orange, and the effect of restoring states after MLP layers in green.

The figure demonstrates higher $\text{AIE}^\ell_\text{mlp}$ values compared to $\text{AIE}^\ell_\text{attn}$ values in both visual and text tower, with a larger contrast in the visual tower. This implies the change of MLP layers contributes more to the final classification, which further validates our choice in performing selection on the MLP layers.

\section{Visualization for Parameter Selection}\label{appen:selection}
\begin{figure}
    \centering
    \includegraphics[height=21mm]{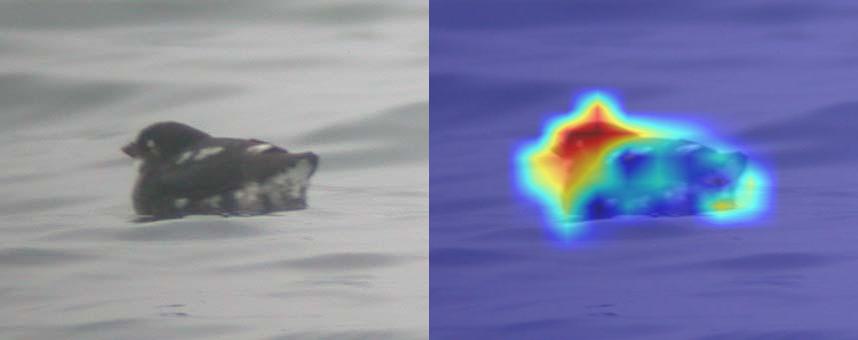}
    \includegraphics[height=21mm]{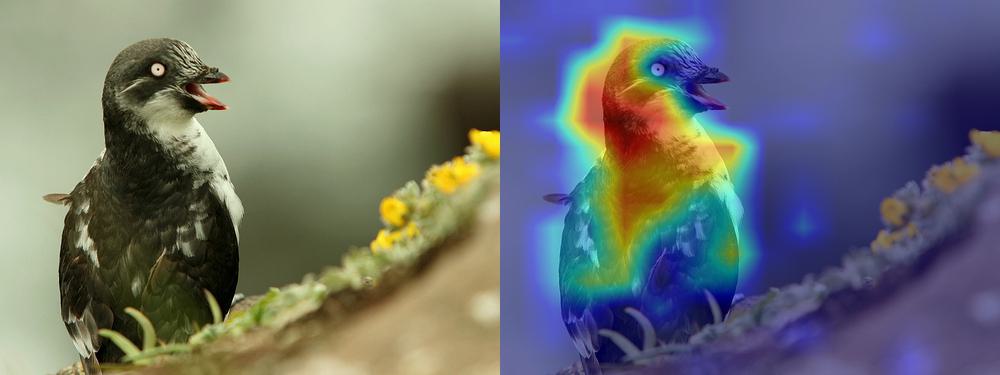}
    \includegraphics[height=21mm]{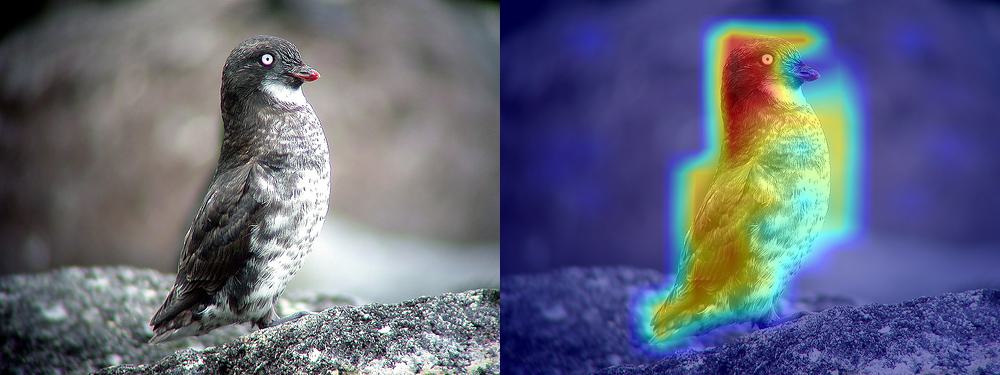}
    
    \includegraphics[height=21.5mm]{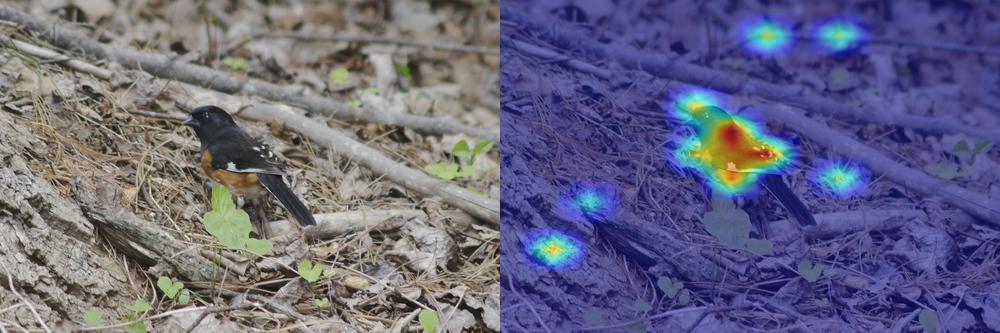}
    \includegraphics[height=21.5mm]{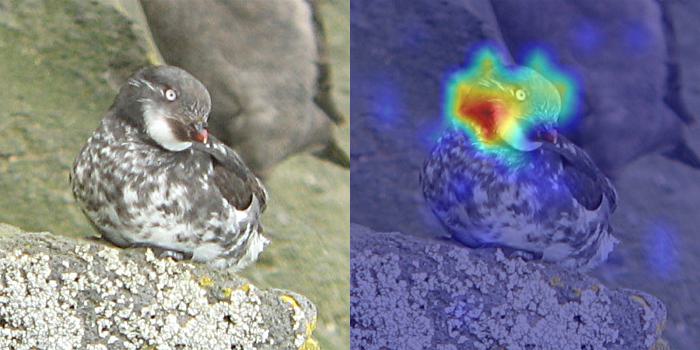}
    \includegraphics[height=21.5mm]{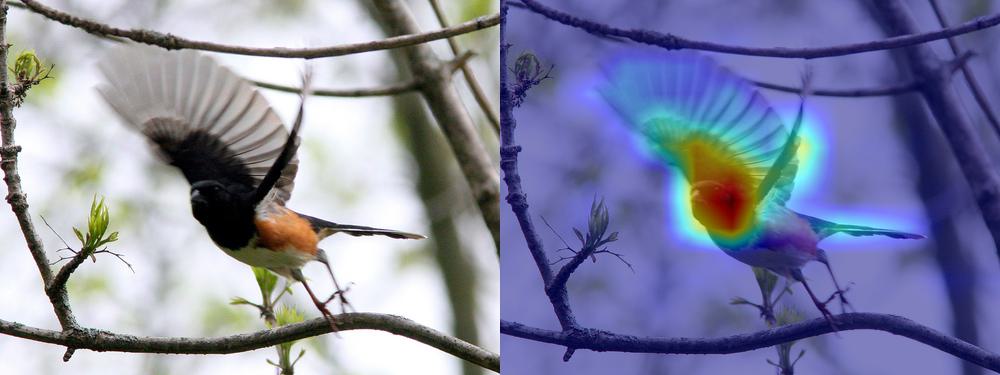}

    \includegraphics[height=28.3mm]{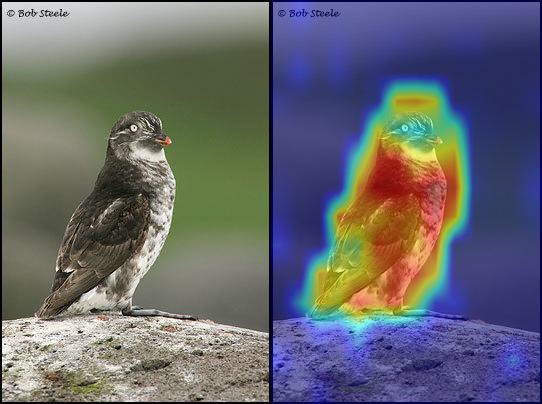}
    \includegraphics[height=28.3mm]{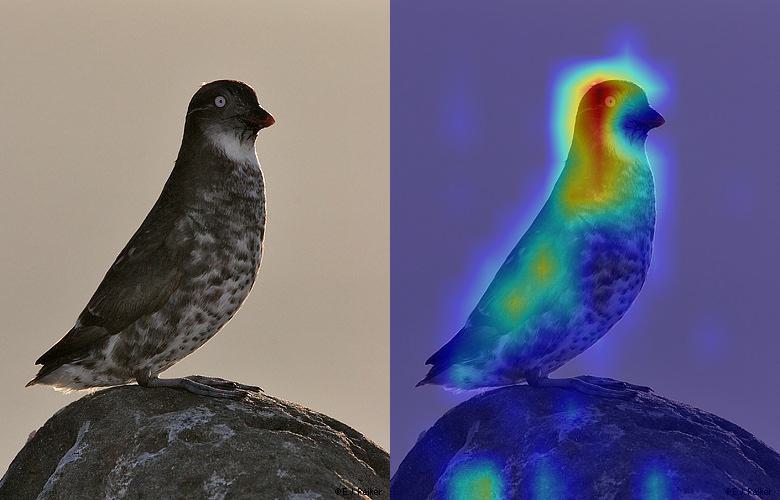}
    \includegraphics[height=28.3mm]{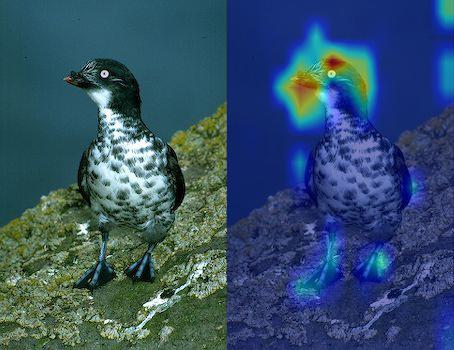}
    \includegraphics[height=28.3mm]{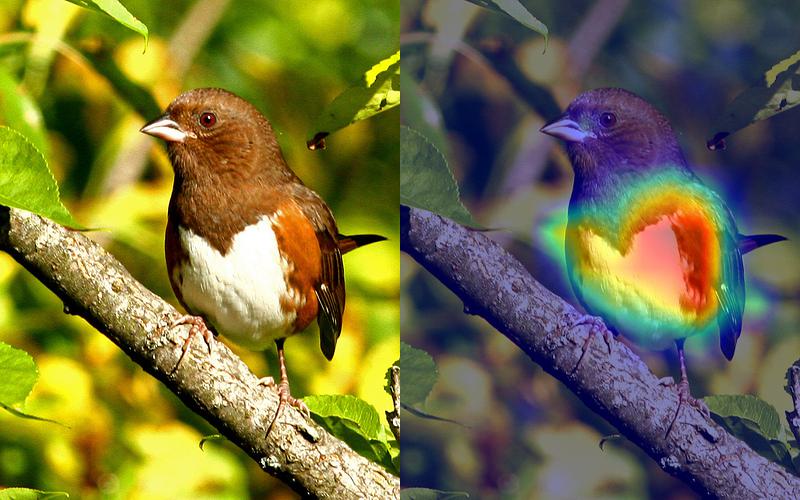}
    \caption{ Highlighted regions by activations of selected neurons in the first MLP layers of the 9th transformer block in gScoreCAM visualization. Selected neurons represent meaningful features in the input image.  }
    \label{fig:camvis}
\end{figure}

In addition to section 5.3, we further validate the parameter selection qualitatively from two perspectives. Firstly, we utilize gScoreCAM~\citep{chen2022gscorecam} to visualize the attention of selected neurons on original images, illustrating their representativeness to the features. Secondly, we visualize the correlation of selected weights from different tasks. This is to demonstrate the task-wise separation in the selection process, which aids in mitigating forgetting.

gScoreCAM~\citep{chen2022gscorecam} follows the idea of ScoreCAM~\citep{wang2020score} to perturb the input image with the upsampled activation map, and aggregate the CAM scores. The importance of neuron activations to specific input features is derived from the aggregated scores. gScoreCAM selects only 10\% of the activations in regard to their gradient values to perturb the input image, and shows the selected activations are effective in localizing the features. This is in agreement with our selection strategy and modularity hypothesis (section 4). We applied gScoreCAM to perform the visualization on the neurons of the first MLP layers of the 9th transformer layers. We selected the top 10\% activation values to perturb the input image. The highlighted regions by selected activations of images from the CUB dataset in shown \cref{fig:camvis}. We    

\begin{figure}
    \centering
    \includegraphics[width=0.35\textwidth]{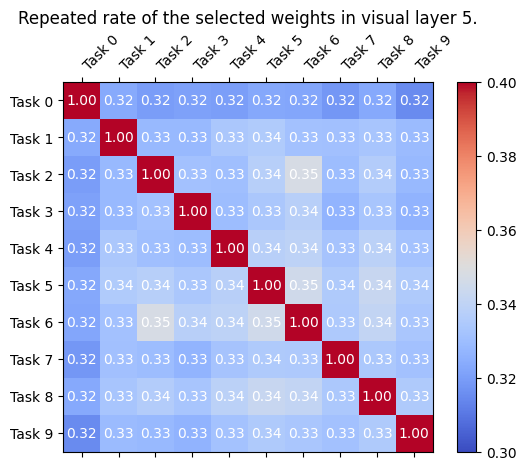}
    \includegraphics[width=0.35\textwidth]{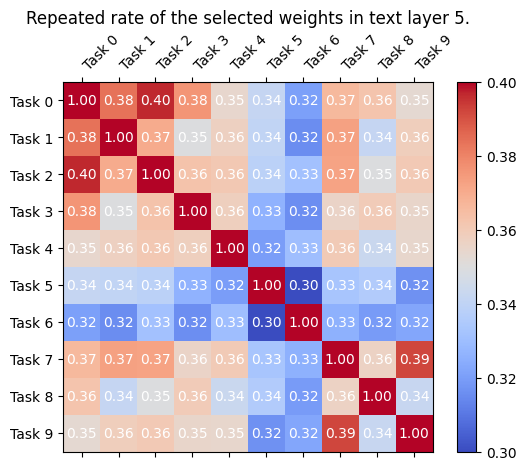}
    
    \includegraphics[width=0.35\textwidth]{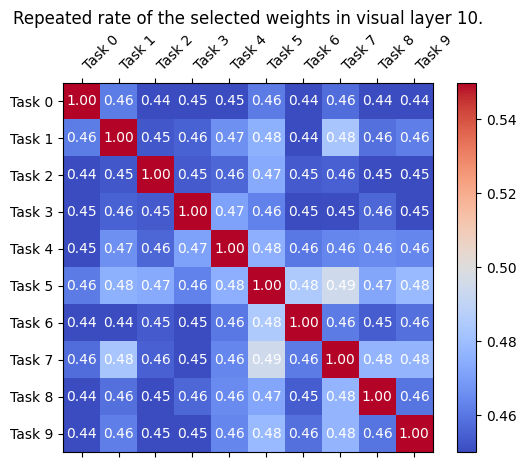}
    \includegraphics[width=0.35\textwidth]{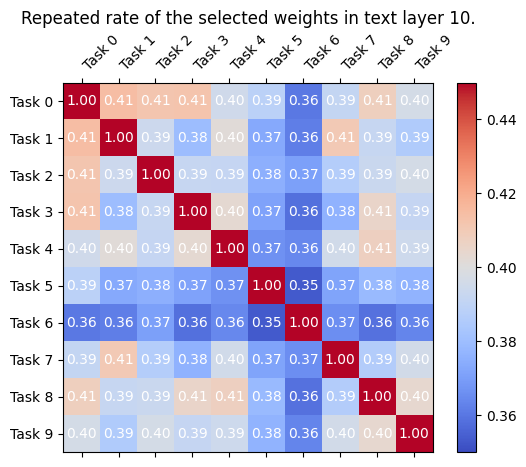}
    \caption{Repeat rate of the selected weight in visual and text tower of layer 5 and layer 10 in CLIP. The shared weight selected two different tasks only counts a small amount of total selected weight. }
    \label{fig:repeat}
\end{figure}

In \cref{fig:repeat}, we analyze the correlation between selected weights across different tasks in various layers. Patch with row label task $i$ and column label task $j$ shows the percentage of the shared weights selected in task $i$ and task $j$. 
We observe that the frequency of repeated weight selection for different tasks seldom exceeds 50\%. This pattern suggests that while our scoring function occasionally identifies common weights across tasks, it predominantly selects task-specific weights.  Notably, the incidence of repeated selection decreases in shallower layers, as demonstrated in layer 5 (first row), which implies a higher occurrence of modulation in these layers.  Thus, our approach of selecting weights across all transformer layers is further validated.

\section{Learnable Scoring Function} \label{sec:learnablescore}
In section 5.3, we perform different scoring functions to validate the effectiveness of our proposed gradient-based scoring function, including the Mask baseline. Here, we describe the Mask baseline.

Although the gradient is an efficient approximation of the parameters' relevance to the task at hand, we suspect selecting parameters independently based on their gradient magnitude might not consider the contribution of the parameters together when updated, and can potentially cause redundancy in the selection. To explore this, we propose to involve an optional optimization stage to adjust the scoring function based on the initial gradient values. Specifically, for parameters $\mathbf{\theta}^l \in \mathbb{R}^{m\times n}$, we define  $\mS\in \mathbb{R}^{m\times n}$ to be the learnable parameters scores. We initialize $\mS$ with the  gradients  computed on the current task, where $\mS_{ij} = \frac{1}{N_t'}\sum_{k=1}^{N_t'} g_{ij} (x_k)$, and consider the estimated gradient as the basis for a target update of the model parameters %
and construct an imaginary  update:
\begin{equation}\label{eq:suppimaginary}
    \theta^{l\prime} =\theta^{l}  -\mu \cdot \mS ,
\end{equation}
where $\mu$ is the update step size (learning rate).
We then optimize $\mS$ by minimizing the task loss $\mathcal{L}$ and an additional  $L_1$ loss ($\|\mS\|_1$)
\begin{equation}
    \mS^\prime= \argmin_{\mS}\mathcal{L}(\theta^{l\prime}; D^t ) + \lambda\|\mS\|_1 ,
    \vspace{-1mm}
\end{equation}
where $\lambda$ is a hyperparameter that weighs the contribution of $L_1$ loss. 
$L_1$ loss is introduced to encourage sparsity in the estimated scores, guiding the optimization to tolerate parameters with large gradient magnitude (and hence large initial scores) when proven relevant to the minimization of the task loss while zeroing out gradients of irrelevant or redundant parameters. %

We optimize $\mS$ for a few epochs. %
 Then, we define $\mathcal{S}(\theta^{l}_{ij}, D^t)=\mS^\prime_{i,j}$ and select top $r$ parameters as the most relevant parameters for the task at hand.  Note that here we estimate parameter scores for one selected layer $\theta^l$, but the formulation can generalize to an arbitrary number of layers.
 
The learnable scoring function requires more computation due to the additional optimization phase of $\mS$ compared to the gradient scores. %
We present the efficacy of this optional stage in Table 3 of the main paper.

\section{Relations to other sparse update works}\label{appen:relation}
\textbf{Gradient-based selection.}
Gradient-based parameter attribution  is  a robust metric that has been widely used in CL and other fields like model compression and multitask learning; however, when and where to use it is the key point to stand our method out.  HAT~\citep{serra2018overcoming}, SupSup~\citep{wortsman2020supermasks}, and SPG~\citep{konishi2023parameter} leverage  the \textit{gradients after training} of previous tasks to penalize the change of previously learned important parameters, which are more similar to EWC~\citep{ewc} and MAS~\citep{mas}.
We applied SPG in our setting and show significant improvements by our method in all metrics in \cref{tab:cilresults}.
More importantly, in our case, gradients of previous data (pre-training data) are inefficient or unavailable to obtain  and noisy. Distinctly, we achieve knowledge  retention via sparse updates by two key steps, localization and parameter \textit{pre-selection}. 
Our gradient-based pre-selection serves as an approximation to identify the specialized  parameters of the upcoming task, which is crucial to allow large decrease in the loss function with the smallest  change in the selected parameters. 

\textbf{Sparse network.}
PiggyBack~\citep{piggyback} learns a task-specific binary mask for every task, requiring task identifiers which we assume inaccessible. SparseCL~\citep{wang2022sparcl} compresses the model by 75\%-95\% for on-device CL, and would fail to fit in our problem in two ways. 
1) During training of a task, SparseCL combines the magnitude of the parameters and their gradients as a score to omit unimportant parameters in Equation (1).
The magnitude of the parameter $\|w\|_1$ is relevant when training a network from scratch and in SparseCL it is the dominant  factor in parameter selection; after reproducing SparseCL experiment, we found that  the  magnitude of weights $\|w\|_1$ is on average 6.93e-3 while the gradient $\alpha\|\frac{\partial\Tilde{\mathcal{L}}(\mathcal{D}_t;\theta)}{\partial w}\|_1$ is on average 2.02e-4. However, in a pre-trained foundation model, the magnitude of weights is mostly relevant to the knowledge learned during pre-training. Our gradient-based selection is to measure the relevance of parameters to the \textit{upcoming task}, and the sparse update is not to compress the network but to discourage the unrelated parameters to be modified. 2) SparseCL dynamically selects parameters to be updated every several epochs;  selecting additional parameters  and omitting from  already changed parameters leads, at the end of the task training,  to many parameters changed. This would incur more forgetting of generic knowledge.
In \cref{tab:cilresults} SparseCL fails to improve Acc. and causes drop in generic knowledge (C.) specially when learning  generic datasets.

\section{Implementation Details}\label{appen:implementation}
\subsection{Dataset}
Here are the statistics of our six experimental benchmarks. 

\textit{Birdsnap}~\citep{birdsnap} Birdsnap is a large bird dataset originally consisting of 49,829 images from 500 bird species with 47,386 images used for training and 2,443 images used for testing. We download the dataset from the official link, and follow the official train-test split. We use a fixed buffer of size 1,500 for this dataset.

\textit{CUB-200-2011}~\citep{cub} The Caltech-UCSD Birds-200-2011 (CUB-200-2011) dataset is for fine-grained visual categorization task. It contains 11,788 images of 200 subcategories belonging to birds, 5,994 for training and 5,794 for testing. We use the Hugging Face implementation of the dataloader.  We use a fixed buffer of size 240 for this dataset.

\textit{CIFAR100}~\citep{cifar100} This dataset has 100 classes containing 600 images each. There are 500 training images and 100 testing images per class. We use the PyTorch implementation of the dataloader. We used a fixed buffer of 2,000 for this dataset. 

\textit{FGVC-Aircraft}~\citep{aircraft} The dataset contains 10,200 images of aircraft, with 100 images for each of 102 different aircraft model variants, most of which are airplanes. The data is divided into three equally sized training, validation, and test subsets. We use the PyTorch implementation of the dataloader, where train and valid set are used for training, and the test set is used for testing.  We use a fixed buffer of size 250 for this dataset.

\textit{Stanford Cars}~\citep{cars} The Stanford Cars dataset contains 16,185 images of 196 classes of cars. The data is split into 8,144 training images and 8,041 testing images, where each class has been split roughly in a 50-50 split. Classes are typically at the level of Make, Model, Year, \eg, 2012 Tesla Model S or 2012 BMW M3 coupe. We use the Hugging Face implementation of the dataloader.  We use a fixed buffer of size 240 for this dataset.

\textit{GTSRB}~\citep{gtsrb} This dataset is designed for recognition of traffic signs. By the time we download it, it contains 43 classes with 26,640 training samples and  12,630 testing samples.  We use the PyTorch implementation of the dataloader. We used a fixed buffer of 1,000 for this dataset. 

For each dataset, during the training, we use the prompt \texttt{a photo of \{\}} with class name as text inputs. We evaluate each baseline on the test set using the original prompts and ensembling strategy provided by~\citet{clip}. 

\subsection{Hyperparameters}

For our algorithm, we use PyTorch implemented AdamW optimizer~\citep{adamw} and learning rate scheduler of Cosine Annealing with Warmup~\citep{loshchilov2016sgdr} for our algorithm, as well as FLYP combined with ER and other CL regularization methods. We use a learning rate of 7.5e-6 and train for 10 epochs for all datasets. The results are reported based on an average of 5 different random seeds. We run all our experiments on one single Nvidia A100 GPU. 

\subsection{Baseline Details}
Here are the implementations for other baselines in Table 1.  

\textbf{FLYP}~\citep{goyal2023finetune} For all FLYP based baselines, we tuned the learning rate in [2.5e-6, 5e-6, 7.5e-6] and training epochs in [5,10,15] and report the best results

\textbf{FLYP+ER}~\citep{er} For ER-based baselines, we apply balanced sampling, where at each step, we sample a balanced batch, half from the current task and half from the previous tasks. 

\textbf{FLYP + MAS}~\citep{mas} We follow the avalanche~\citep{avalanche} to implement  MAS regularizer with  FLYP. To cope with the  large-scale architecture, we normalize the estimated weights' importance   by their maximum value. We tuned the scaling factor of MAS loss in [0.01, 0.05, 0.1] and report the best results.

\textbf{FLYP + ER + LwF/PRD }~\citep{lwf,prd} For LwF, we follow the implementation of avalanche. For PRD, we follow the official implementation. We further  tuned temperature in [0.01,0.1,1.0,5.0] and loss scaling factor in [0.01, 0.05, 0.1] and report the best results.

\textbf{L2P, DualPrompt}~\citep{l2p,dualprompt} These two methods were originally designed for ViT backbone with linear classifier. They proposed to freeze the feature extractor and only train the classifier. We adopt the idea to the backbone of CLIP architecture with pre-trained weights provided by \texttt{timm} library, where we freeze the visual and text feature extractors and only train the linear projection layers. We applied the prompt techniques on the visual tower of CLIP. 
A class balanced buffer is applied to them as what we did for FLYP + ER. These methods are highly tailored for ImageNet pre-trained transformers and do not scale to other pre-trained weights, leading to surprisingly bad performance when combined with CLIP, in spite of our best efforts to tune the hyperparameters carefully.  

\textbf{SLCA}~\citep{slca} We adopted the slow learning rate and classifier alignment to the CLIP backbone. In the first training phase, we applied the learning of $1.5e-6$ to the backbone, and $7.5e-6$ to the projection layers. In the section training phase of classifier alignment, we only train the projection layers. 

\textbf{LoRA-EWC}~\citep{loraewc} We compute fisher information by  CC12m~\citep{changpinyo2021cc12m}, and apply the EWC loss on every task. We applied the LoRA architecture on both visual and text tower. We further modified it with different ranks in LoRA and applied a replay buffer.

\noindent Here are other baselines in Table 2.

\textbf{Random} This baseline mainly follows our method SPU, except for Equation 2 in the main paper.  In this baseline, we use random values for the scoring function. 

\textbf{Mask} We described this method in \cref{sec:learnablescore} in the supplement. We optimize the learnable score matrix $\mS$ for 5 epochs, with the learning rate of $5e-4$ and step size $\mu=5e-4$. We set the $L_1$ loss coefficient $\lambda=1e-3$

\textbf{PiggyBack}~\citep{piggyback}  This baseline mainly follows the Mask baseline, except for the format of the imaginary update in \cref{sec:learnablescore}. We applied the PiggyBack mask learning format, where 
\begin{equation}
    \theta^{l\prime} =\theta^{l}  + \mu \cdot \text{m}(\mS) .
\end{equation}
Here $\mu$ is a scaling factor, and $\text{m}(\cdot)$ is a binary mask. We uniformly initialized $\mS$ and applied AdamW optimizer as proposed in PiggyBack. We optimize the score matrix $\mS$ for 5 epochs, with the learning rate of 1e-4 and the scaling factor $\mu$ of 1e-5.

\section{More Details in Ablation Study}\label{appen:fullresults}

 \paragraph{Knowledge Retention.}
 \begin{table}[h]
\centering
\resizebox{\textwidth}{!}{%
\begin{tabular}{l|ccc|ccc|ccc|ccc|ccc|ccc|ccc}
\toprule
 & \multicolumn{3}{c}{Aircraft} & \multicolumn{3}{c}{Birdsnap} & \multicolumn{3}{c}{Cars} & \multicolumn{3}{c}{CIFAR100} & \multicolumn{3}{c}{CUB} & \multicolumn{3}{c}{GTSRB} &  \multicolumn{3}{c}{Average} \\ \midrule
 & Acc. & F. & C. & Acc. & F. & C. & Acc. & F. & C. & Acc. & F. & C. & Acc. & F. & C. & Acc. & F. & C.  & Acc. In.  & Avg. F. & C. Drop \\ \midrule
 \rowcolor{lightgray}
w/ Loc. & 44.43 & 14.42 & 63.48 & 55.35 & 12.78 & 61.94 & 77.51 & 3.26 & 63.42 & 83.99 & -0.39 & 61.38 & 71.51 & 4.84 & 62.87 & 94.25 & -7.87 & 62.55  &  \textbf{21.34}&\textbf{4.51}& \textbf{0.94}\\
w/o Loc. & 43.35& 16.12& 63.54& 54.56& 14.37& 61.08& 76.83& 4.30& 63.26& 84.43& -0.26& 59.95& 71.64& 4.56& 62.42& 93.81& -7.75& 61.53& 20.93& 5.22&1.59\\ \bottomrule
\end{tabular}%
}
\caption{Comparison between w/ localization and w/o localization. The localization improves the retention ability.  }\label{tab:localization}
\end{table}

 Our good retention ability is dually contributed by the localization and selective update. We localize the change to the first MLP layer, and keep  other model components unchanged for knowledge retention. In~\cref{tab:localization}, selective update without localization (w/o Loc.) results in less retention.  Besides localization's role in retention and the sparse updates, the parameters randomly selected generally have smaller gradients magnitudes regarding the task-in-hand; 
thus under same learning rate and number of epochs, the magnitude of parameters change  can be  smaller, which helps in retention. 

\paragraph{Neuron-based Selection.} 

\begin{table}[h]
\centering
\resizebox{\textwidth}{!}{%
\begin{tabular}{l|ccc|ccc|ccc|ccc|ccc|ccc|ccc}
\toprule
 & \multicolumn{3}{c}{Aircraft} & \multicolumn{3}{c}{Birdsnap} & \multicolumn{3}{c}{Cars} & \multicolumn{3}{c}{CIFAR100} & \multicolumn{3}{c}{CUB} & \multicolumn{3}{c}{GTSRB} &  \multicolumn{3}{c}{Average} \\ \midrule
 & Acc. & F. & C. & Acc. & F. & C. & Acc. & F. & C. & Acc. & F. & C. & Acc. & F. & C. & Acc. & F. & C.  & Acc. In.  & Avg. F. & C. Drop \\ \midrule
 \rowcolor{lightgray}
Weight & 44.43 & 14.42 & 63.48 & 55.35 & 12.78 & 61.94 & 77.51 & 3.26 & 63.42 & 83.99 & -0.39 & 61.38 & 71.51 & 4.84 & 62.87 & 94.25 & -7.87 & 62.55  &  \textbf{21.34}&4.51& 0.94\\
Neuron & 44.13 & 14.02 & 63.60 & 55.32 & 12.66 & 62.77 & 77.46 & 3.33 & 63.62 & 83.98 & -0.79 & 61.84 & 71.14 & 5.16 & 63.23 & 93.54 & -8.15 & 63.22  &  21.09&\textbf{4.37}& \textbf{0.50}\\ \bottomrule
\end{tabular}%
}
\caption{Comparison between weight-based selection and neuron-based selection. Our method employs weight selection and has better learning ability.  }\label{tab:fullselectionstrategy}
\end{table}

We propose to compute the element-wise importance scores by Equation 2 in the main paper to facilitate weight-based selection. 
Whereas,~\citet{aljundi2018selfless} put forth a technique to calculate row-wise importance scores to perform neuron-based selection. 

\cref{tab:fullselectionstrategy} shows the full results of variants of  selection strategy, where the gray row represents our strategy. In the baseline named ``Weight'' we compute an element-wise scoring function by Equation 2, and select the top 10\% entries of each weight matrix to update. In the baseline named ``Neuron'', we compute a row-wise scoring function based on the row summation of the element-wise scoring function by Equation 2. Then we select the 10\% rows of each weight matrix to update. 

We find that weight-based selection yields slightly improved learning performance while exhibiting a marginal decrease in hold-out accuracy. Nonetheless, the overall performance trends remain comparable between the two strategies. This observation highlights the robustness of our localization and importance scoring methods to any of the selection strategy.

\paragraph{Selection Rate.} In Section 5.3, we present the average results  of our method  under varying selection rates. \cref{tab:ablationratefull} shows the full results, where the gray row represents our reported results. Our main results select the top 10\% elements localized layer. We compare to the baselines where the top 1\% or the top 50\% are selected for update. All other configurations are kept the same. 

\begin{table}[h!]
\resizebox{\textwidth}{!}{%
\begin{tabular}{l|ccc|ccc|ccc|ccc|ccc|ccc|ccc}
\toprule
 & \multicolumn{3}{c}{Aircraft} & \multicolumn{3}{c}{Birdsnap} & \multicolumn{3}{c}{Cars} & \multicolumn{3}{c}{CIFAR100} & \multicolumn{3}{c}{CUB} & \multicolumn{3}{c}{GTSRB} & \multicolumn{3}{c}{Average}\\ \midrule
 & Acc. & F. & C. & Acc. & F. & C. & Acc. & F. & C. & Acc. & F. & C. & Acc. & F. & C. & Acc. & F. & C.  & Acc. In.  & Avg. F. &C. Drop \\ \midrule
{0.01} & 37.64 & 11.45 & 63.54 & 53.49 & 9.87 & 62.25 & 74.62 & 2.20 & 63.40 & 83.79 & -1.64 & 60.91 & 66.79 & 4.39 & 63.01 & 88.89 & -7.71 & 61.53  & 17.70& \textbf{3.10}&1.11\\
 \rowcolor{lightgray}
{0.10} & 44.43 & 14.42 & 63.48 & 55.35 & 12.78 & 61.94 & 77.51 & 3.26 & 63.42 & 83.99 & -0.39 & 61.38 & 71.51 & 4.84 & 62.87 & 94.25 & -7.87 & 62.55  & 21.34& 4.51&\textbf{0.94}\\
{0.50} & 46.73 & 20.74 & 63.56 & 53.96 & 17.98 & 61.72 & 77.64 & 6.12 & 63.48 & 83.47 & 1.51 & 61.53 & 71.89 & 8.06 & 62.85 & 95.74 & -7.85 & 62.43  & \textbf{21.73}& 7.76&0.95\\ \bottomrule
\end{tabular}%
}
\centering\caption{Full results of ablation on selection rate. Our method select 10\% weights, achieving better trade-off in learning and forgetting. }\label{tab:ablationratefull}
\end{table}

\paragraph{Buffer Size.}  In Section 5.3, we present the average results  of our method and FLYP + ER  under varying buffer size.
We study  buffer sizes of 1\%, 2\% and 4\% of the total dataset size. 
\cref{tab:fullbuffersize} shows the full results of buffer size ablation. We report our method with 4\%  buffer size  of the total dataset size in Table 1 in the main paper, highlighted in gray.  
\begin{table}[h!]
\resizebox{\textwidth}{!}{%
\begin{tabular}{cc|ccc|ccc|ccc|ccc|ccc|ccc|ccc}
\toprule
\multirow{2}{*}{Method} & \multirow{2}{*}{\makecell{Buffer Size\\ / Total Size }} & \multicolumn{3}{c}{Aircraft} & \multicolumn{3}{c}{Birdsnap} & \multicolumn{3}{c}{Cars} & \multicolumn{3}{c}{CIFAR100} & \multicolumn{3}{c}{CUB} & \multicolumn{3}{c}{GTSRB} & \multicolumn{3}{c}{Average} \\ \cmidrule{3-23} 
 &  & Acc. & F. & C. & Acc. & F. & C. & Acc. & F. & C. & Acc. & F. & C. & Acc. & F. & C. & Acc. & F. & C.  &  Acc. In.  & Avg. F. &C. Drop\\ \midrule
ER & 1\% & 27.76 & 44.49 & 49.22 & 43.76 & 33.59 & 55.90 & 61.22 & 21.19 & 55.34 & 73.70 & 13.34 & 40.14 & 53.39 & 25.26 & 48.81 & 93.03 & -4.22 & 16.79  & 8.97& 22.27&19.18\\
ER & 2\%  & 33.42 & 41.37 & 49.74 & 49.96 & 29.20 & 56.35 & 62.83 & 21.45 & 57.35 & 78.72 & 7.94 & 41.74 & 57.90 & 22.84 & 50.83 & 95.64 & -6.68 & 15.82  & 13.24& 19.35&18.24\\
 \rowcolor{lightgray}
ER & 4\%  & 41.42 & 31.48 & 50.41 & 56.22 & 21.63 & 56.72 & 69.08 & 16.42 & 58.07 & 82.86 & 3.41 & 42.10 & 64.07 & 17.72 & 51.30 & 96.28 & -7.40 & 17.34  & 18.48& 13.88&17.56\\ \midrule
SPU & 1\%  & 37.82 & 21.96 & 63.56 & 47.54 & 23.61 & 61.65 & 73.68 & 6.84 & 63.36 & 80.44 & 4.87 & 61.49 & 66.06 & 9.32 & 62.47 & 90.55 & -4.93 & 62.76  & 16.18& 10.28&1.00\\
SPU & 2\%  & 40.65 & 20.31 & 63.44 & 51.33 & 18.97 & 61.90 & 75.00 & 6.17 & 63.41 & 82.45 & 2.03 & 61.36 & 68.39 & 8.42 & 62.81 & 92.99 & -7.04 & 62.63  & 18.63& 8.14&0.96\\
 \rowcolor{lightgray}
SPU & 4\%  & 44.43 & 14.42 & 63.48 & 55.35 & 12.78 & 61.94 & 77.51 & 3.26 & 63.42 & 83.99 & -0.39 & 61.38 & 71.51 & 4.84 & 62.87 & 94.25 & -7.87 & 62.55  & 21.34& 4.51&0.94\\ \bottomrule
\end{tabular}%
}
\centering\caption{Full results of ablation on the buffer size. Our method shows superior performance over ER even in smaller buffer scenarios. }\label{tab:fullbuffersize}
\end{table}

\paragraph{Task Length}
\begin{table}[h!]
\resizebox{\textwidth}{!}{%
\begin{tabular}{cc|ccc|ccc|ccc|ccc|ccc|ccc|ccc}
\toprule
\multirow{2}{*}{Method} & \multirow{2}{*}{\makecell{Task\\ Length }} & \multicolumn{3}{c}{Aircraft} & \multicolumn{3}{c}{Birdsnap} & \multicolumn{3}{c}{Cars} & \multicolumn{3}{c}{CIFAR100} & \multicolumn{3}{c}{CUB} & \multicolumn{3}{c}{GTSRB} & \multicolumn{3}{c}{Average} \\ \cmidrule{3-23} 
 &  & Acc. & F. & C. & Acc. & F. & C. & Acc. & F. & C. & Acc. & F. & C. & Acc. & F. & C. & Acc. & F. & C.  &  Acc. In.  & Avg. F. &C. Drop\\ \midrule
ZSCL & 10 & 30.96 & 15.65 & \textbf{65.53} & 49.85 & 13.28 & 63.13 & 67.79 & 8.27 & 62.90 & 80.50 & 1.05 & \textbf{61.90} & 61.09 & 7.69 & 62.78 & 62.92 & 13.54 & {62.92} & 9.01 & 9.91 & \textbf{0.36}\\
ER & 10  & 41.42 & 31.48 & 50.41 & {56.22} & 21.63 & 56.72 & 69.08 & 16.42 & 58.07 & 82.86 & 3.41 & 42.10 & 64.07 & 17.72 & 51.30 & {96.28} & -7.40 & 17.34 & 18.48 & 13.88 & 17.56\\
 \rowcolor{lightgray}
SPU & 10  & 44.43 & 14.42 & 63.48 & 55.35 & 12.78 & 61.94 & 77.51 & 3.26 & 63.42 & 83.99 & -0.39 & 61.38 & 71.51 & 4.84 & 62.87 & 94.25 & -7.87 & 62.55  & 21.34& 4.51&0.94\\ \midrule
ZSCL & 20  &  28.23& 28.81& 62.92& 43.23& 20.23& 62.83& 69.67& 11.21& 62.56& 68.05& 21.21& 55.17& 60.55& 16.15& 62.15& 15.40& 33.43& 55.82& -2.32& 21.84& 3.31\\
ER & 20  & 5.67& 37.93& 47.99& 53.53& 28.12& 55.80& 65.71& 22.60& 52.98& 81.73& 9.74& 32.58& 61.58& 23.25& 47.21& 94.80& -0.33& 9.72& 15.66& 20.22& 22.50\\
 \rowcolor{lightgray}
SPU & 20  &39.60& 13.95& 63.70& 54.52& 13.12& 62.41& 75.13& 6.40& 63.01& 83.77& 3.43& 61.33& 68.57& 8.32& 62.78& 92.53& -2.37& 62.27& 19.18 & 7.14&0.97 \\ \bottomrule
\end{tabular}%
}
\centering\caption{Full results of ablation on the task length. Our method shows superior performance over ER even in longer task scenarios. }\label{tab:fulllength}
\end{table}

We perform experiments on 20-split datasets and compare our method with ER (second-best Acc. In.) and ZSCL (best C.) in \cref{tab:fulllength}. The gap between SPU and ER/ZSCL becomes larger, as shown in blue  value, than that in 10-split experiments. With 20 tasks, SPU   has almost no drop in performance compared to 10 tasks, while ER and ZSCL have negative overall performance (Acc. In. - C. Drop).

\section{More Details in Efficiency}\label{appen:efficiency}

\begin{table}[h]
\resizebox{\textwidth}{!}{%
\begin{tabular}{c|ccc|ccc|ccc|ccc|ccc|ccc|ccc}
\toprule
\multicolumn{1}{l}{} & \multicolumn{3}{c}{Aircraft} & \multicolumn{3}{c}{Birdsnap} & \multicolumn{3}{c}{Cars} & \multicolumn{3}{c}{CIFAR100} & \multicolumn{3}{c}{CUB} & \multicolumn{3}{c}{GTSRB} & \multicolumn{3}{c}{Average}\\ \cmidrule{2-22} 
 & Acc. & F. & C. & Acc. & F. & C. & Acc. & F. & C. & Acc. & F. & C. & Acc. & F. & C. & Acc. & F. & C.  & Acc. In.  & Avg. F. &C. Drop\\ \midrule
one batch & 44.42 & 14.40 & 63.50 & 55.02 & 13.33 & 62.04 & 77.41 & 3.36 & 63.40 & 83.99 & -0.38 & 61.36 & 71.48 & 4.90 & 62.86 & 94.14 & -7.79 & 62.52  & 21.24 & 4.64 &\textbf{0.94} \\
 \rowcolor{lightgray}
0.25 & 44.43 & 14.42 & 63.48 & 55.35 & 12.78 & 61.94 & 77.51 & 3.26 & 63.42 & 83.99 & -0.39 & 61.38 & 71.51 & 4.84 & 62.87 & 94.25 & -7.87 & 62.55  & 21.34 & 4.51 &\textbf{0.94} \\
0.50 & 44.33 & 14.48 & 63.48 & 55.31 & 12.73 & 61.88 & 77.54 & 3.16 & 63.44 & 84.03 & -0.41 & 61.37 & 71.67 & 4.63 & 62.87 & 94.24 & -7.82 & 62.58  & 21.35& 4.46 &\textbf{0.94} \\
1.00 & 44.40 & 14.40 & 63.47 & 55.28 & 12.61 & 61.85 & 77.61 & 3.12 & 63.44 & 84.05 & -0.40 & 61.35 & 71.66 & 4.64 & 62.87 & 94.27 & -7.81 & 62.58  & \textbf{21.37}& \textbf{4.43} &0.96 \\ \bottomrule
\end{tabular}%
}
\centering \caption{Full results of ablation on the number of samples to compute the gradient approximation. Our scoring function can efficiently cope with only one-batch gradient accumulation. } \label{tab:fullgradient}
\end{table}

\paragraph{Number of samples to compute gradient approximation.}
In Equation 2, we accumulate the gradients of  $N'_t$ samples to approximate the importance. Here we ablate the effect of accumulating gradients with one batch (128 data points), 25\% 50\%, and 100\% of the current set. We compute the importance score by the accumulated gradients before the training of every task, and the computational cost per task gets reduced with fewer samples to approximate the scoring function. With more samples, the accuracy is slightly increased, with also slight decrease in forgetting. Our algorithm is robust to all different configurations in general. Full results are shown in \cref{tab:fullgradient}. We choose to report our main results by accumulate gradients of 25\% samples of the current set, highlighted in gray.

\end{document}